\documentclass[sigplan,screen, balance=false]{acmart}
\usepackage{soul}
\usepackage{caption}
\newcommand{\captionFontSize}{footnotesize}
\AtBeginDocument{%
  \providecommand\BibTeX{{%
    \normalfont B\kern-0.5em{\scshape i\kern-0.25em b}\kern-0.8em\TeX}}}

\setcopyright{acmcopyright}
\copyrightyear{2023}
\acmYear{2023}
\acmDOI{XXXXXXX.XXXXXXX}

\acmConference[AI-2023]{AI-2023 Forty-third SGAI International Conference on Artificial Intelligence}{Dec 12--14,
  2023}{Cambridge, England}
\begin{document}

\title[CorrEmbed: Evaluating Pre-trained Model Efficacy]{CorrEmbed: Evaluating Pre-trained Model Image Similarity Efficacy with a Novel Metric}

\author{Karl Audun Kagnes Borgersen}
\affiliation{%
  \institution{University of Agder}
  \streetaddress{Jon Lilletuns vei 9}
  \city{Grimstad}
  \country{Norway}}
\email{karl.audun.borgersen@uia.no}

\author{Morten Goodwin}
\affiliation{%
  \institution{University of Agder}
  \streetaddress{Jon Lilletuns vei 9}
  \city{Grimstad}
  \country{Norway}}
\email{morten.goodwin@uia.no}

\author{Jivitesh Sharma}
\affiliation{%
  \institution{University of Agder}
  \streetaddress{Jon Lilletuns vei 9}
  \city{Grimstad}
  \country{Norway}}
\email{jivitest.sharma@uia.no}

\author{Tobias Aasmoe}
\affiliation{%
  \institution{Tise}
  \streetaddress{Mariboes gate 8}
  \city{Oslo}
  \country{Norway}}
\email{tobias@tise.com}

\author{Mari Leonhardsen}
\affiliation{%
  \institution{Tise}
  \streetaddress{Mariboes gate 8}
  \city{Oslo}
  \country{Norway}}
\email{mari.leonhardsen@tise.com}

\author{Gro Herredsvela Rørvik}
\affiliation{%
  \institution{FJONG}
  \streetaddress{Lilletorget 1}
  \city{Oslo}
  \country{Norway}}
\email{gro@fjong.com}

\renewcommand{\shortauthors}{Borgersen, et al.}

\begin{abstract}
  Detecting visually similar images is a particularly useful attribute to look to when calculating product recommendations. Embedding similarity, which utilizes pre-trained computer vision models to extract high-level image features, has demonstrated remarkable efficacy in identifying images with similar compositions. However, there is a lack of methods for evaluating the embeddings generated by these models, as conventional loss and performance metrics do not adequately capture their performance in image similarity search tasks.

  In this paper, we evaluate the viability of the image embeddings from numerous pre-trained computer vision models using a novel approach named CorrEmbed. Our approach computes the correlation between distances in image embeddings and distances in human-generated tag vectors. We extensively evaluate numerous pre-trained Torchvision models using this metric, revealing an intuitive relationship of linear scaling between ImageNet1k accuracy scores and tag-correlation scores. Importantly, our method also identifies deviations from this pattern, providing insights into how different models capture high-level image features.

  By offering a robust performance evaluation of these pre-trained models, CorrEmbed serves as a valuable tool for researchers and practitioners seeking to develop effective, data-driven approaches to similar item recommendations in fashion retail.\footnote{All code and experiments are openly available at \url{https://anonymous.4open.science/r/CorrEmbedRecSys-731F}}
\end{abstract}

\ccsdesc[300]{Computing methodologies~Image representations}
\ccsdesc[300]{Computing methodologies~Neural networks}

\keywords{Image Similarity, Content-Based Recommendations, Zero-Shot Learning, Recommender Systems}

\begin{teaserfigure}
  \includegraphics[width=\textwidth]{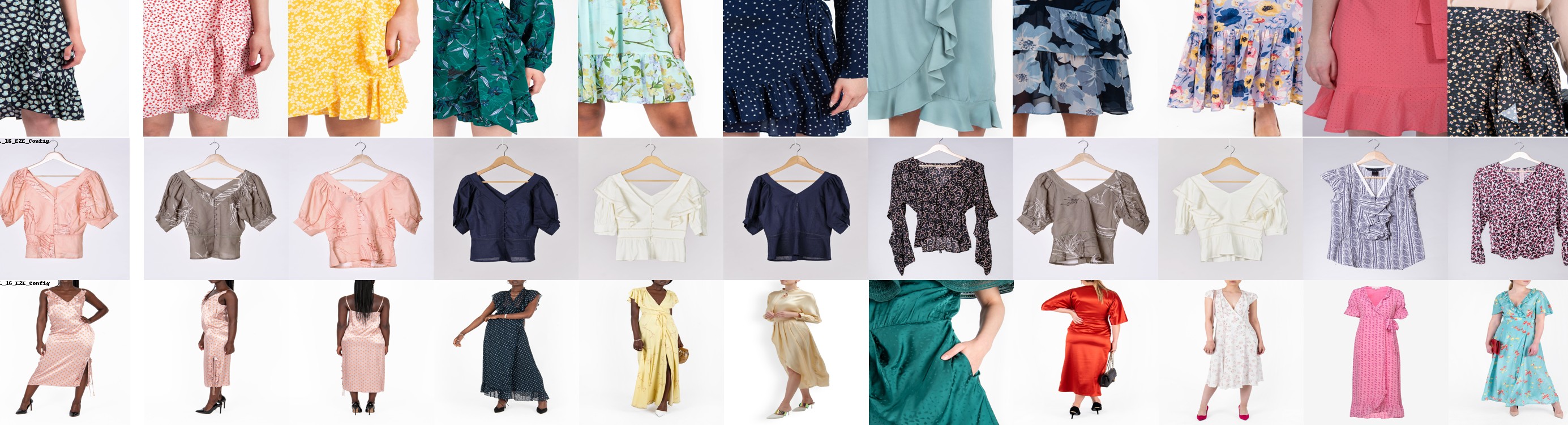}
  \caption{Demonstration of image embedding search. The original image is displayed to the far left, with the most similar images being displayed beside it.}
  \Description{Vision transformers based image similarity}
  \label{fig:teaser}
\end{teaserfigure}


\maketitle

\section{Introduction}

There are several enticing aspects to using pre-trained computer vision models to recommend similar products. It requires no resources for model training, avoiding the need for labeled data or the computation required to train machine learning models. Neither does it require any product information beyond a product image, lending itself particularly well to relatively small online storefronts or online second-hand sales. Similar item recommendations entail simply recommending items similar to a target item. Fashion, in particular, as a domain has some distinctive properties that make it uniquely suited for recommendations based on image similarity search, namely its emphasis on the visual appearance of the products. Extracting image embeddings from pre-trained models for Recommender Systems(RS) is used in production today. Tise\footnote{Tise: \url{tise.com}}, for instance, is a large Norwegian second-hand sales company and the employer of two of our co-authors. The company used embedding comparisons in production as part of an ensemble similar-item recommendation method.

However, existing methods face validation challenges because direct comparisons of image embeddings do not align with the intended use case of pre-trained image models. Consequently, neither their loss nor classification performance metrics effectively indicate a model's performance in this domain. While the efficacy of retrieving similar images based on computer vision model embeddings is apparent to human observers, limited literature exists on evaluating their effectiveness. This paper contributes a more rigorous evaluation of how well each of the models performs as compared to human tag annotation.


This paper introduces CorrEmbed, an evaluation metric based on tag-based similarity within the fashion domain. It takes advantage of human-tagged outfits to evaluate the zero-shot performance of a model's embeddings. The evaluation is performed by calculating the correlation between distances in image-embedding space and in tag-embedding space.  These indicator variables, or tag vectors, are augmented by weighting them according to category using statistical entropy. 

CorrEmbed is used to provide a benchmark for the performance of numerous pre-trained computer-vision models. We discuss the performance of these and which features and model architectures are more conducive to good tag-correlation performance, such as the format of the tensors produced by different pre-trained model versions. In particular, we underline cases in which this performance deviates from the pattern of increased image-classification performance leading to increased CorrEmbed performance.

The pre-trained models evaluated for this project are from Torchvision's model set\cite{torcvisionmodels}.

\section{Related Work}
RS can refer to numerous different approaches, from Collaborative Filtering\cite{recsys_cf_advances_2021} to Deep Learning\cite{recsys_dl_survey_2021}\cite{recsys_dl_survey_2020} to (Deep) Reinforcement Learning\cite{recsys_rl_survey_2022} to Tsetlin Machines\cite{recsys_tm_2023}.
At the core of this paper lies the concept of similar item recommendations. While collaborative filtering models have long been the dominant method for providing recommendations, content-based RSs have often been employed to address problems with linear scaling. These item-based RSs identify relationships between items and recommend items frequently bought together, for which the capability of extracting information from for example text descriptions or images is highly relevant\cite{item_based_recs}. Image embeddings have also been used in the context of classification for a while. Akata et al.\cite{zero_shot_embeddings} use it to perform zero-shot learning on unsupervised data by detecting clusters of image embeddings. Fu. et al. \cite{zero_shot_survey_2019} provide an overview of zero-shot learning in 2019. These papers demonstrate how prevalent the use of image embeddings is in the SotA.

Visual image attributes form the foundation of numerous contemporary RSs, such as \cite{visual_style_learning} and \cite{image_rec_hamming}. Evaluating how similar images are is a necessity within Image Retrieval, in this context, Tarasov et al.\cite{image_similarity_search} utilize the embeddings of a trained neural network for this task for which the distance between two images is the Euclidian distances between embeddings. Garcia et al.\cite{non_metric_image_similarity} train a regression NN for this purpose and compare performance to a few other metrics, including cosine distance. Resnick et al. \cite{user_similarity_correlation} utilize Pearson correlation to measure user similarity. While these previous works parse image (or user) embeddings in a manner comparable to CorrEmbed, none of them quantitatively evaluate how well individual models perform at measuring image similarity.

\section{Methods} \label{chap:methods}
In summary, CorrEmbed entails retrieving image embeddings from pre-trained classification computer vision models and identifying similar items by calculating the distance between them. We evaluate their performance by calculating the correlation of the distances between pairs of image embeddings and the distances between pairs of tag embeddings. The final score represents the mean correlation between image and tag embeddings across $k$ samples.

The datasets used in this paper are generated using data from FJONG, a small clothing rental company with access to approximately $10,000$ human-tagged outfits and around $18,000$ corresponding images. $705$ tags constitute the tag-embedding space, each belonging to one out of the $13$ tag categories listed in Table \ref{tab:fjong_tags}.

All models and model weights are retrieved from Torchvision's model set\footnote{TorchVision's model set is available at \url{https://pytorch.org/vision/stable/models.html}}. All classification models have trained on the ImageNet\cite{imagenet1k} dataset with $1k$ classes.

\subsection{Tag-Based metric}
A one-hot-encoded vector of tags is calculated based on tag presence, resulting in a vector with $705$ dimensions. Similarly to the image-embeddings, the clothing items are converted into tag-based representations. We calculate the distances between an input embedding $i$ and all $n$ other embeddings for both tag and image embeddings. (Eq.~\ref{eq:cosine_sim},\ref{eq:tag_similarity},\ref{eq:image_similarity}) in which $T$ and $I$ refer to a set of all image and tag embeddings respectively and $T_i$ or $I_i$ represents the ith element in both of these sets. For a given set of image embeddings, we assess the performance by calculating the correlation between the tag-based metric distance and the distance between image embeddings. This is done using the Pearson correlation coefficient. The final score for a model is obtained by computing the mean correlation across $k$ image samples (Eq.~\ref{eq:corrembed}). In which $x_i$ and $y_i$ represent a set of $n$ tag and image similarity scores for the vector at index $i$ and $\bar{x}$ and $\bar{y}$ are the mean values of the same sets.

\begin{equation}
\text{cosine\_similarity}(\mathbf{A}, \mathbf{B}) = \frac{\mathbf{A} \cdot \mathbf{B}}{\|\mathbf{A}\| \|\mathbf{B}\|}
\label{eq:cosine_sim}
\end{equation}

\begin{equation}
\small\text{Image\_Similarity}(I_i) = y_i = \{ \text{cosine\_similarity}(\mathbf{I}_i, \mathbf{I}_j) : j = 1, 2, \dots, n \}
\label{eq:image_similarity}
\end{equation}

\begin{equation}
\small\text{Tag\_Similarity}(T_i) = x_i = \{ \text{cosine similarity}(\mathbf{T}_i, \mathbf{T}_j) : j = 1, 2, \dots, n \}
\label{eq:tag_similarity}
\end{equation}

\begin{equation}
CorrEmbed = \frac{1}{k}\sum_{i=1}^{k}\frac{\sum_{j=1}^{n} (x_{ji} - \bar{x})(y_{ji} - \bar{y})}{\sqrt{\sum_{j=1}^{n} (x_{ji} - \bar{x})^2}\sqrt{\sum_{j=1}^{n} (y_{ji} - \bar{y})^2}}
\label{eq:corrembed}
\end{equation}

In this dataset, tags are grouped into categories, e.g., ``Collar" is categorized as ``Neckline". All tags present in the dataset, except for brand tags, are shown in Table \ref{tab:fjong_tags}. As the ``Size" and ``Shoe Size" category isn't necessarily present in the images of the dataset (The same clothing item in different sizes will occasionally share the same product photo), this category is dropped entirely.
For the context of recommendation, some tags are more compelling to the average customer than others. We are more interested in representations that appropriately capture the user's interests. A user browsing for a new winter coat will be more interested in other winter coats rather than products with the same color. We evaluate the customers' purchase history and compute the entropy for each tag category. Calculating the entropy was done to capture the likelihood of a customer re-purchasing an item with a similar category, yielding a lower value if the purchased outfits' tag categories exhibit consistent sub-tags. For example, if a user exclusively buys clothing items with ``Dots" or ``Stripes" patterns, the ``Pattern" category will have a low entropy score. Conversely if the user prefers a variety of colors, the ``Color" category will have a high entropy score. As shown in Equation \ref{eq:entropy}, in which $C_i$ refers to a customer's rental history, c is the total number of customers evaluated, $C_i(x_j)$ is the total occurrences of tag $x_j$ in the rental history $C_i$, and $X$ refers to a tag category for which we want to calculate the weights.

\begin{equation}
\text{Entropy(X)} = H(X) = \frac{1}{c}\sum_{i=1}^{c}-\sum_{j=1}^{|X|} \frac{C_i(x_j)}{\|C_i\|} \log \frac{C_i(x_j)}{\|C_i\|}
\label{eq:entropy}
\end{equation}

We normalize these entropy values between $0$ and $1$ based on the maximum possible entropy within this range, and subsequently invert the weights (Eq.~\ref{eq:tag_weights}), $min(H)$ and $max(H)$ refer to the largest and smallest calculated tag category value. Our earlier tag embeddings are then weighted for their respective tag category. This ensures, for example, that the relative tag distance between a blue blazer and a red blazer is shorter than the distance between a blue blazer and a blue jumpsuit.

\begin{equation}
\text{Tag\_Weights}(X) = 1 - \frac{H(X) - \text{min}(H)}{\text{max}(H) - \text{min}(H)}
\label{eq:tag_weights}
\end{equation}

\begin{table*}[ht]
\centering
\captionsetup{font=\captionFontSize}
\resizebox{0.85\textwidth}{!}{%
\begin{tabular}{|
p{0.18\linewidth} |
p{1\linewidth} |}
\hline
Category & Tags \\ \hline
Brand & \textbf{552 different fashion brands, omitted for brevity.} \\ \hline
Material & Triacetate, Lyocell, Polyester, Cashmere, Linen, Cupro, Velvet, Leather, Spandex, Lace, Beaded, Faux fur, Fur, Rayon, Down, Acrylic, Bamboo, Polyethylene Terephthalate, Acetate, Satin, Chiffon, Silk, Polyamide, Tulle, Wool, Nylon, Denim, Cotton, Vegan Leather, Viscose, Modal, Gold, Elastane, Lurex, Lycra, Lacquer, Silver, Tencel, Polyvinyl Chloride, Polyurethane, Metal\\ \hline
Category & Blouses, Accessories, Jewelry, Shirts, Kimonos, Shoes, Bags, Pants, Suits, Coats, Sweaters, Vests, Outerwear, Jumpsuits, Skirts, Blazers, Dresses, Cardigans, Knitwear, Jackets, Shorts, Tops\\ \hline
Color & Yellow, Purple, Grey, Blue, Green, Brown, Pink, Multicolor, Beige, Orange, Black, Gold, Red, White, Navy, Silver, Turquoise, Burgundy\\ \hline
Size & XXS, Onesize, Large, 4XL, Medium, Extra Small, 3XL-4XL, Extra Large, 3XL, Small, 2XL, XXL-XXXXL\\ \hline
No category & Sporty, Winter, Height - 180-190 cm, Dressed-up, Fall, New, Summer, Romantic, Edgy, Spring, Classic\\ \hline
Occasion & FJONG Active, Going out, Black-tie, Everyday, FJONG Plus Size, Prom, Wedding, Party, FJONG Bump, Active, Business\\ \hline
Sleeve & Mid arms, Spaghetti straps, T-shirt, Cold shoulder, Tube, Straps, Long arms\\ \hline
Embellishment & Ruffles, Pearls, Sequins, Feathers, Glitter, Studs, Tassels\\ \hline
Neckline & Boat Neck, Deep Neck, Halter Neck, V-neck, Round Neck, Collar, Turtleneck\\ \hline
Waist & Empire, High, Normal, Adjustable, Stretchy, Low\\ \hline
Shoe Size & 39, 36, 40, 38, 37, 41\\ \hline
Pattern & Floral, Checkers, Dots, Stripes, Animal, Pattern\\ \hline
Fit & Loose fit, Wrap, Pregnant-friendly, Slim fit\\ \hline
Length & Midi, One, Mini, Maxi\\ \hline
\end{tabular}}
\caption{Tags present in the dataset of this paper}
\label{tab:fjong_tags}
\end{table*}

\begin{figure}[ht]
  \centering
  \includegraphics[width=0.45\textwidth]{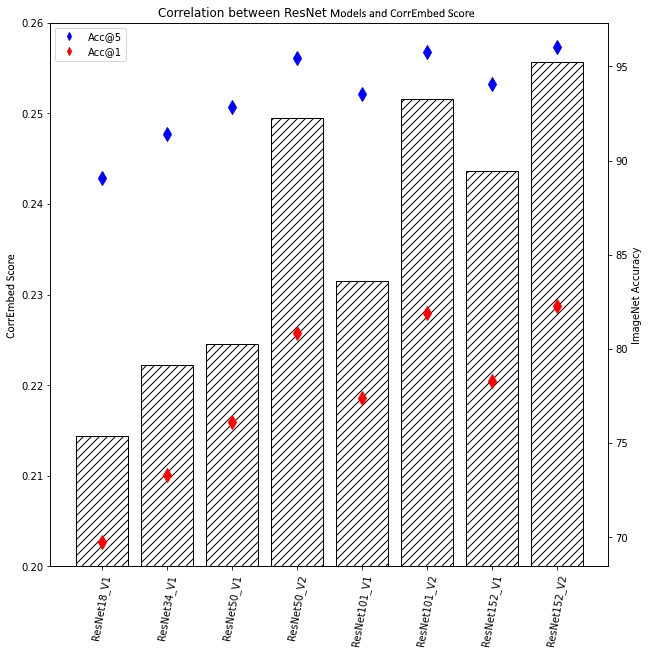}
  \caption{Comparison between ResNet \cite{resnet_paper} model correlation score and ImageNet1k accuracy across differing ResNet model sizes. CorrEmbed scores have been constrained to the range of $0.2$ to $0.26$ to emphasize the score differences.}
  \label{fig:resnet_performance_comparison}
\end{figure}

\section{Results and Discussions}
This section presents the results of our experiments, which benchmark the performance of our pre-trained models. We also employ t-Distributed Stochastic Embedding (t-SNE) plots to visualize the clustering of embeddings in our models.

Model performance on CorrEmbed generally increases by the model size and ImageNet performance, as seen in Figure \ref{fig:resnet_performance_comparison}. While model performance doesn't necessarily directly scale in accordance with its ImageNet score, scaled-up versions of the same model outperform their smaller counterparts. This provides a good sanity check on the veracity of our metric. 
The correlation between the ImageNet score and the CorrEmbed score holds true for comparisons between the scaled-up versions of the same architecture but isn't necessarily the case when comparing two different model architectures. For instance, EfficientNet models outperform ResNet models with the same accuracy score, e.g., ResNet50 V2 has an Acc@1 score of $80.858$ and achieves a CorrEmbed score of $0.249$, as compared to EfficientNet B2 which has an Acc@1 score of $80.608$ and a CorrEmbed score of $0.273$.

\begin{figure}[ht]
  \centering
  \includegraphics[width=0.5\textwidth]{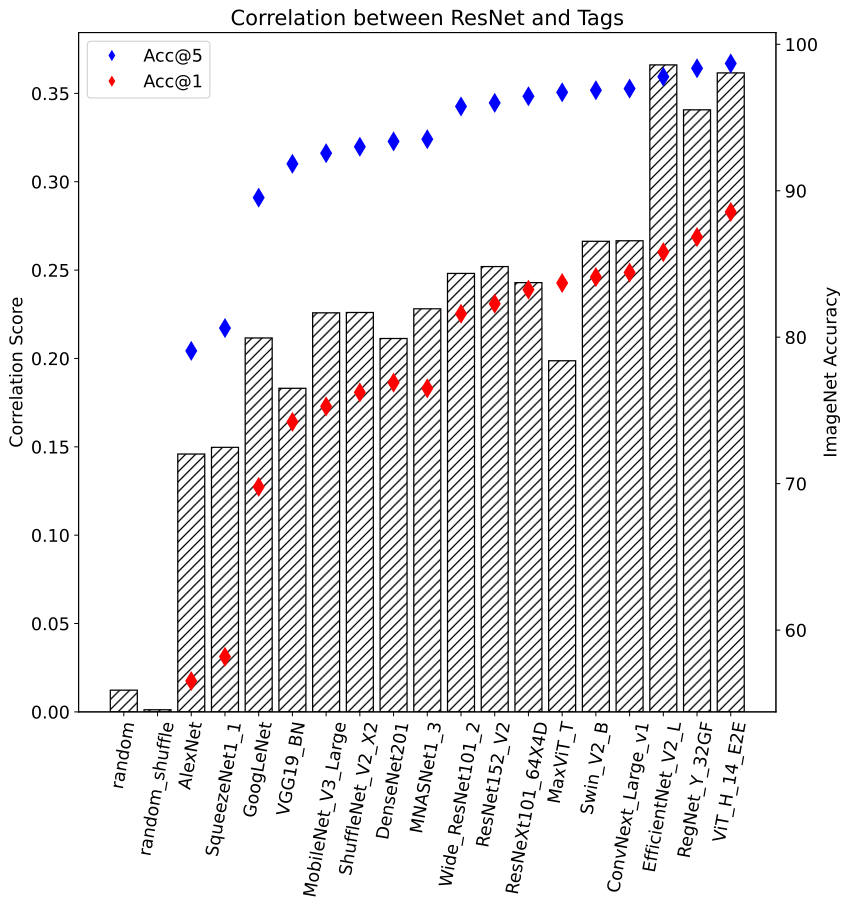}
  \caption{Graph representation of the  results from Table \ref{tab:results_overview}. Columns display the CorrEmbed score for the different models. The blue and red lines show performance on ImageNet}
  \label{fig:results_overview}
\end{figure}

Table \ref{tab:results_overview} (and Figure \ref{fig:results_overview}) provides an overview of the performance of various model architectures. We selected the top-performing version of each architecture based on ImageNet1k accuracy. We include two control methods for both tag and image embeddings to establish a baseline for significance. The first random model method generates a random tensor with the same shape as the image embeddings for each image, while the random tag vectors generate a binary vector with the same shape as the tag vector. The two shuffle control methods randomize the association between tag or image embeddings and each outfit.


We observe a significant correlation between ImageNet1k scores and CorrEmbed scores, albeit with some deviations. The top-performing models ViT, RegNetY\cite{regnetxy}, and EfficientNet\cite{efficientnet_v2} outperform other models by a margin comparable to the performance improvements between AlexNet\cite{alexnet} and ConvNext\cite{convnext}, despite negligible differences in ImageNet1k scores. This observation makes some intuitive sense, as improving a model's accuracy score from, for instance, $87$ to $88$ likely corresponds to a much greater enhancement of the model's internal representation than an improvement from $56$ to $57$, in line with the Pareto Principle\footnote{\url{https://en.wikipedia.org/wiki/Pareto_principle}}. However, the observed increase in performance exceeds what we would expect if this were the sole contributing factor.

The embeddings evaluated in this case were all retrieved as the standard output embeddings of the models. This had the added advantage of removing the shape of the embeddings as a factor for the CorrEmbed score since the final output embedding of the classes remains the same $1000$-dimensional vector regardless of the model architecture.

\begin{table*}[ht]
\centering
\captionsetup{font=\captionFontSize}
\resizebox{0.8\textwidth}{!}{%
\begin{tabular}{
|l|llllll| }
 \hline Model& ImgNet Acc@1 & ImgNet Acc@5 & CorrEmbed & Unweighted & Random & Shuffled \\ \hline\hline
random & 0.0 & 0.0 & 0.0086 & 0.0088 & 0.0229 & 0.008 \\ \hline
random shuffle & 0.0 & 0.0 & 0.0016 & -0.0032 & 0.0076 & -0.0058 \\ \hline
AlexNet & 56.522 & 79.066 & 0.1463 & 0.1485 & 0.0165 & 0.0008 \\ \hline
SqueezeNet1 1 & 58.178 & 80.624 & 0.1508 & 0.137 & 0.012 & 0.0038 \\ \hline
GoogLeNet & 69.778 & 89.53 & 0.2103 & 0.1933 & 0.0133 & 0.0036 \\ \hline
VGG19 BN & 74.218 & 91.842 & 0.1837 & 0.1733 & 0.0129 & 0.0023 \\ \hline
MobileNet V3 Large & 75.274 & 92.566 & 0.2251 & 0.201 & 0.0174 & 0.0048 \\ \hline
ShuffleNet V2 X2 & 76.23 & 93.006 & 0.2264 & 0.1977 & 0.0175 & 0.0051 \\ \hline
DenseNet201 & 76.896 & 93.37 & 0.2121 & 0.1903 & 0.0146 & 0.0023 \\ \hline
MNASNet1 3 & 76.506 & 93.522 & 0.2276 & 0.2083 & 0.0148 & 0.0058 \\ \hline
Wide ResNet101 2 & 81.602 & 95.758 & 0.2445 & 0.2182 & 0.0075 & 0.0034 \\ \hline
resnet152 V2 & 82.284 & 96.002 & 0.2493 & 0.2243 & 0.0119 & 0.0043 \\ \hline
ResNeXt101 64X4D & 83.246 & 96.454 & 0.2393 & 0.2092 & 0.0096 & 0.0048 \\ \hline
MaxViT T & 83.7 & 96.722 & 0.1943 & 0.174 & 0.0044 & 0.0039 \\ \hline
Swin V2 B & 84.112 & 96.864 & 0.2649 & 0.2417 & 0.0151 & 0.0045 \\ \hline
ConvNext Large v1 & 84.414 & 96.976 & 0.2641 & 0.2383 & 0.0157 & 0.0032 \\ \hline
EfficientNet V2 L & 85.808 & 97.788 & \textbf{0.3680} & \textbf{0.3189} & 0.0176 & 0.0051 \\ \hline
RegNet Y 32GF & 86.838 & 98.362 & 0.3428 & 0.306 & 0.014 & 0.0038 \\ \hline
ViT H 14 E2E & \textbf{88.552} & \textbf{98.694} & 0.3633 & 0.3184 & 0.0114 & 0.0036 \\ \hline
\end{tabular}}
\caption{Overview of the performance using the top-performing version of each model architecture, sorted by ImageNet1k score. ImgNet Acc@1 and ImgNet Acc@5 refer to the model's performance on accuracy at ImageNet1k top 1 and top 5, respectively. CorrEmbed is our weighted tag vector scoring, and unweighted refers to the same vectors without the weights. Random and shuffle in both rows and columns are control methods. The CorrEmbed scores of this table are visually represented in Figure \ref{fig:results_overview}}
\label{tab:results_overview}
\end{table*}

The layers preceding the output layer tend to capture a finer representation of the embeddings. Though the degree to which this is the case depends on the model itself. Table \ref{tab:results_overview_final_input} details the scores of the models based on the input embeddings to the penultimate layer rather than the output. Working with embeddings earlier than this is unfortunately too inconsistent to make any direct comparisons. As happened with SqueezeNet\cite{SqueezeNet} in Table \ref{tab:results_overview_final_input}, any embeddings retrieved before the model pools into the shape $(1, -1)$ tend to reach unworkable sizes. 

Table \ref{tab:results_overview_final_input} also logs inference time for batches of $90$ images, along with the shapes of the evaluated tensors. Unsurprisingly, the best-performing models also clocked in the longest inference time, though in contrast to ImageNet accuracy, some models surpassed expectations significantly based solely on inference time. MaxViT\cite{maxvit} and MobileNet\cite{mobilenetv3} are good examples of these. An explanation for this could be the priorities of the original model developers. As we've only evaluated the top-performing models of each architecture, these are likely the model iterations with the heaviest focus on performance to the detriment of other aspects of the model and are, therefore, subject to a significant degree of diminishing returns. Interestingly, despite the varying tensor shapes, the CorrEmbed score is even more closely associated with the ImageNet1k score in the penultimate model layers (Table \ref{tab:results_overview} compared to the output layer in Table \ref{tab:results_overview_final_input}, have a Pearson correlation of $0.767$ and $0.941$ respectively compared to ImageNet $accuracy@1$)

\begin{table*}[ht]
\centering
\captionsetup{font=\captionFontSize}
\resizebox{0.8\textwidth}{!}{%
\begin{tabular}{
|l|llll| }
 \hline Model & Model Params & CorrEmbed & Inference Time & Embedding Shape \\ \hline\hline
    AlexNet & 61.1M & 0.1775 & 0.052 & (90, 4096) \\ \hline
    SqueezeNet1 1 & 1.2M & N/A & 0.094 & (90, 1000, 13, 13) \\ \hline
    GoogLeNet & 6.6M & 0.2167 & 0.064 & (90, 1024) \\ \hline
    VGG19 BN & 143.7M & 0.2247 & 0.158 & (90, 4096) \\ \hline
    MobileNet V3 Large & 5.5M & 0.2299 & 0.054 & (90, 1280) \\ \hline
    ShuffleNet V2 X2 & 7.4M & 0.2290 & 0.057 & (90, 2048) \\ \hline
    DenseNet201 & 20.0M & 0.2406 & 0.151 & (90, 1920) \\ \hline
    MNASNet1 3 & 6.3M & 0.2375 & 0.079 & (90, 1280) \\ \hline
    Wide ResNet101 2 & 68.9M & 0.2462 & 0.264 & (90, 2048) \\ \hline
    resnet152 V2 & 60.2M & 0.2491 & 0.202 & (90, 2048) \\ \hline
    ResNeXt101 64X4D & 83.5M & 0.2469 & 0.608 & (90, 2048) \\ \hline
    MaxViT T & 30.9M & 0.2516 & 0.197 & (90, 512) \\ \hline
    Swin V2 B & 87.9M & 0.2659 & 0.404 & (90, 1024) \\ \hline
    ConvNext Large v1 & 197.8M & 0.2695 & 0.559 & (90, 1536) \\ \hline
    EfficientNet V2 L & 118.5M & 0.3604 & 1.022 & (90, 1280) \\ \hline
    RegNet Y 32GF & 145.0M & 0.3824 & 1.137 & (90, 3712) \\ \hline
    ViT H 14 E2E & 633.5M & \textbf{0.4288} & 15.412 & (90, 1280) \\ \hline
\end{tabular}}
\caption{Overview of CorrEmbed performance based on the input to the last layer. Inference time is the time taken to evaluate $\frac{1}{200}th$ of the dataset ($90$ images). Control values for both image and tag embeddings, along with binary tag vectors have been omitted for brevity. These maintain roughly the same ratio to the CorrEmbed score as seen in Table \ref{tab:results_overview}. As the final layer of SqueezeNet is an Average Pooling layer, we were unable to perform our experiment on it due to the size of the tensors it produced. The inference time documented for ViT was run in a separate batch from the rest of the models shown. It is, therefore, possible external factors have influenced it.}
\label{tab:results_overview_final_input}
\end{table*}

The use of models trained on ImageNet, in particular, could potentially have had the added advantage of the models ignoring any humans appearing in the image. For example, the $4th$ most similar image detected by the upper model in Figure \ref{fig:whiteDressComparison}. As none of ImageNet1k's classes involve classifying humans\cite{imagenet1k}, the pre-trained models are incentivized to ignore them. This congrues well with our metrics emphasizing the properties of worn clothing rather than the people wearing them. The ability to overlook the human subjects in images is more pronounced in the stronger models compared to the weaker ones.

\subsection{Evaluating Fashion-CLIP}
To gain a better idea of how well the ImageNet1k models perform as compared to other models, we run CorrEmbed on Fashion-CLIP\cite{fashion_clip}, an open vocabulary model fine-tuned on the fashion domain. The model achieved a score of $0.396$ as compared to our top-performing model of ViT\_H\_14\_E2E, which achieved a score of $0.471$\footnote{This experiment was performed on a later iteration of the dataset with significantly reduced noise, this results in somewhat inflated CorrEmbed scores as compared to the scores in Table \ref{tab:results_overview}}. Interestingly, the Fashion-CLIP model performs worse than ViT\_H despite being fine-tuned to the relevant domain. 

\subsection{Exploring embedding space}
\begin{figure*}[ht]
  \includegraphics[width=\textwidth]{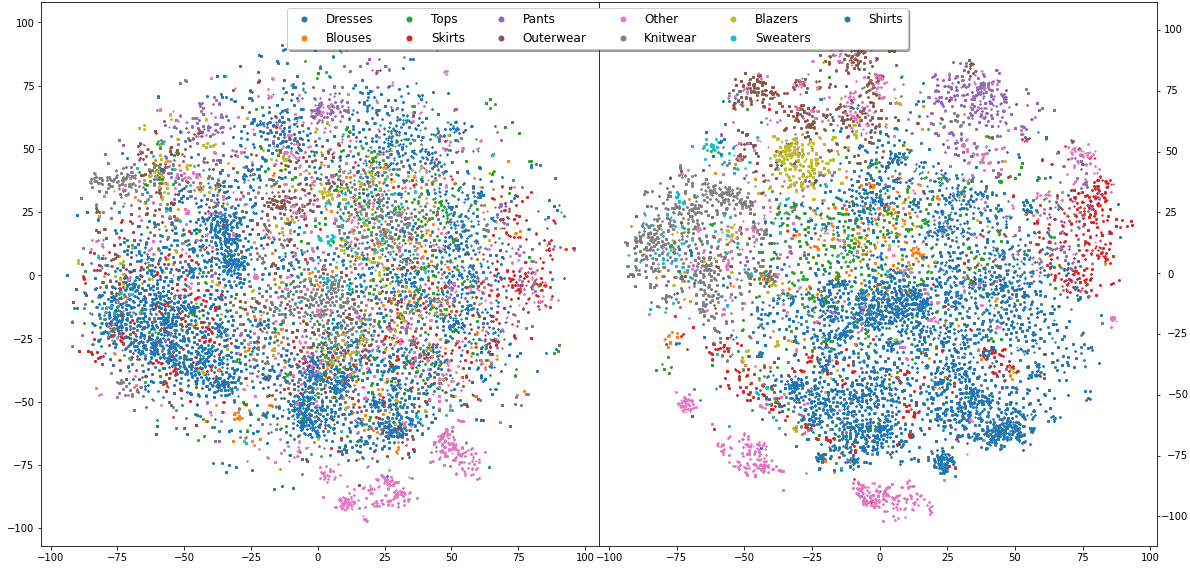}
  \caption{t-SNE diagrams of our dataset's images based on embeddings generated using ResNet50 V1 (left) and ViT H E2E (right). The embeddings are colored according to the ``Category" tag category. Only the top $10$ categories are shown to maintain legibility. All others are relegated to the "Other" category. }
  \label{fig:tsne_collage}
\end{figure*}

As shown in Figure \ref{fig:tsne_collage}, the high-dimensional embeddings of each image can be visualized in low-dimensional space by taking into account the relative distances between each point using t-SNE\cite{tsne_visual}. Despite the models used not being trained in this domain we observe some clear clustering of embeddings based on their tags. Figure \ref{fig:tsne_collage} label embeddings tagged with the same ``category" (essentially, the type of clothing). Moreover, we notice significant improvements in clustering for the better-performing model on the right compared to the model on the left.

An expected flaw with recommending similar items using embeddings is evident in the t-SNE diagrams. Some embeddings are naturally going to end up in isolated positions far away from the larger clusters of embeddings. These isolated items will effectively never be recommended based solely on similar item recommendations. Steps can be taken to increase the recommendation priority of these isolated items, but embeddings placed on the outskirts of clusters will likely rank far too low in the similar item rankings among its neighbors to ever actually be recommended in a natural setting.


\subsection{Dataset Scale}
The dataset used for this paper is relatively small, consisting of around $18000$ images in total. This scale ensures that the direct comparison of all image embeddings is feasible even in production when used for similar item recommendations. However, this format also leads to a significant number of outlier compositions. A good example is the input image in Figure \ref{fig:woodBackgroundComparison}. The top-performing model picks up on the worn fleece sweater and finds other warm clothing items to be the most similar, while the lower-performing models are more affected by composition. This makes some intuitive sense. Each image in the ImageNet1k dataset consists of a central subject to be classified and more-or-less irrelevant background information. The better-performing models are the ones able to adequately capture these subjects while avoiding getting hung up on the other details of the images. It is difficult to tell the degree to which the models are affected by the composition, as the majority of the FJONG dataset consists of images with uniform backgrounds and often without a fashion model. A more expansive dataset would better demonstrate how badly composition impacts similarity predictions.

Further, the tagging of outfits within FJONG is significantly noisier than we would prefer. The process of tagging them has been carried out over several years by several different individuals. Therefore, whether and how certain categories of tags, such as ``Color" or ``Material" has been applied is somewhat inconsistent. While we have not applied CorrEmbed to any other fashion datasets, the noise present in our current one is a good indication of transferability to more thoroughly vetted datasets.

\subsection{Qualitative performance analysis}
\begin{figure*}[ht]
\centering
  \includegraphics[width=\textwidth]{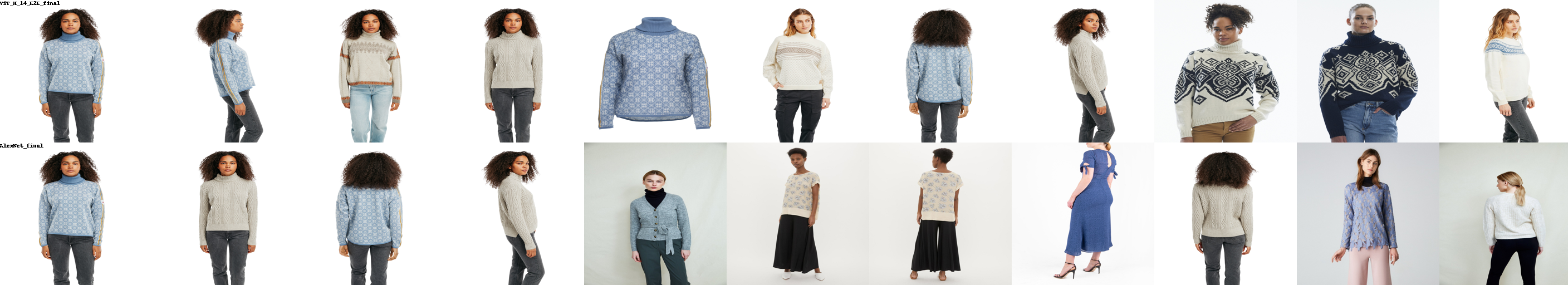}
  \caption{Comparison between a strong and weak model. The original image is displayed to the far left, with the most similar images being displayed in order from left to right. The top row shows the stronger ViT\_H\_14\_E2E model while the row below shows the results from the weaker AlexNet model.}
  \label{fig:whiteDressComparison}
\end{figure*}

\begin{figure*}[ht]
  \includegraphics[width=\textwidth]{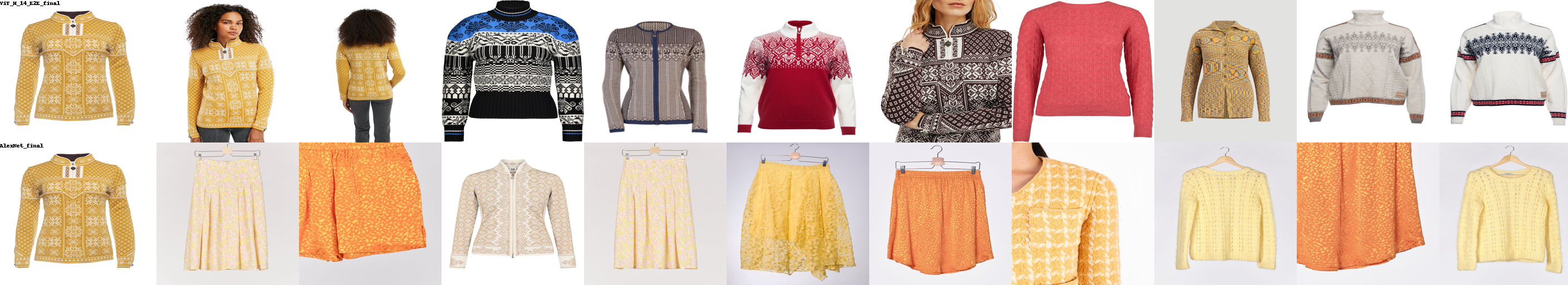}
  \caption{Comparison between a strong and weak model. The original image is displayed to the far left, with the most similar images being displayed in order from left to right. The top row shows the stronger ViT\_H\_14\_E2E model while the row below shows the results from the weaker AlexNet model.}
  \label{fig:sweater_comparison}
\end{figure*}

The main purpose of this paper is to achieve a concrete metric with which the capabilities of embedding evaluation could be measured. Such a metric is of little use, however, if the image similarity does not improve in the eyes of a human as well. This section discusses a few qualitative observations concerning the capabilities of higher-performing methods as compared to lower-performing ones. Figure \ref{fig:sweater_comparison} emphasizes how strong the contrast between the best and worst performing models, ViT\_H\_14\_E2E and AlexNet, can be. The former model can easily pick out a set of similar sweaters, while the latter gets bogged down picking out clothes with similar color schemes. Significant differences exist in the attributes that the models choose to focus on, as illustrated in Figure \ref{fig:whiteDressComparison}. Weaker models tend to put too much emphasis on whether a fashion model is present in the image, as well as the model's pose. Stronger models, on the other hand, are more capable of ignoring these factors and, for instance, picking out the original clothing item worn in the input image.

\begin{figure*}[ht]
  \includegraphics[width=\textwidth]{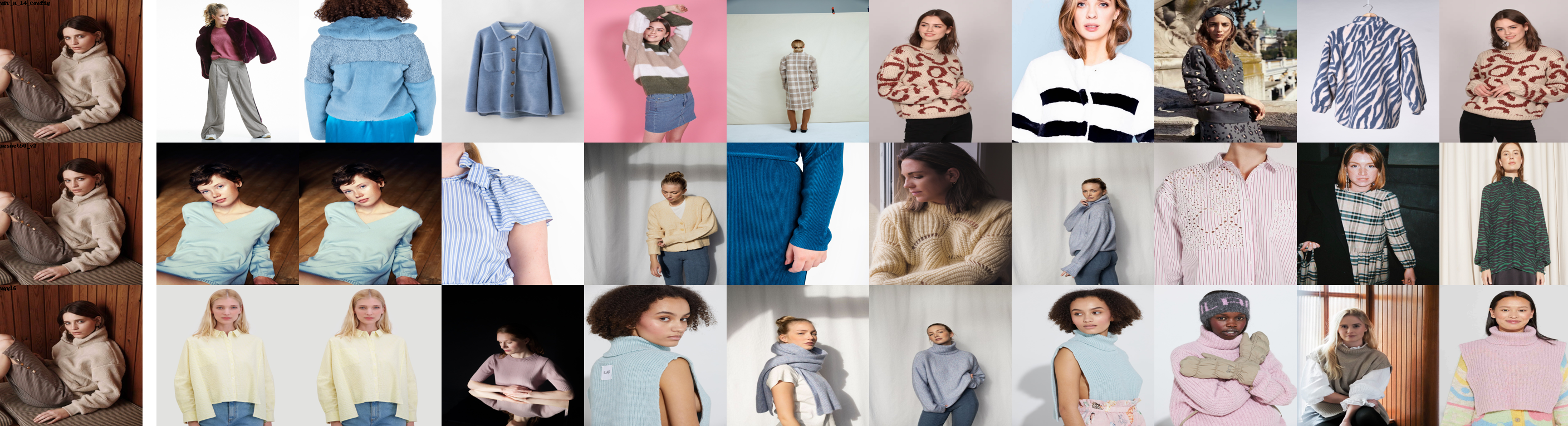}
  \caption{Comparison between ViT\_H (top), resnet50\_v2 (middle), and vgg16 (bottom) with a difficult image. The ViT performs better than the resnet50 model which perform better than the vgg16 model.}
  \label{fig:woodBackgroundComparison}
\end{figure*}

The input image in Figure \ref{fig:woodBackgroundComparison} is among the most challenging images in the dataset. The uncommon composition introduces a degree of visual noise, which causes difficulties with image embedding comparison. The extent to which the visual noise affects the embeddings varies. For example, the products identified by the $vgg16$ model appear close to random, while the outfits found by the $ViT_H$ model focus on the fleece sweater. Note that the actual product sold from the input image is, in reality, the skirt, not the sweater. This highlights another limitation of this method, the inability to specify which component of the image is most relevant.

\section{Conclusion}

In this paper, we introduce and evaluate the performance of CorrEmbed, a novel method designed to assess the zero-shot image-embedding similarity performance of NNs. We employ this metric to evaluate a myriad of different pre-trained ImageNet1k classification models. Of these models, the vision transformer ViT\_H\_14\_E2E is found to be the most performant, beyond even the performance of Fashion-CLIP, a model tuned to the fashion domain. CorrEmbed score is strongly correlated to ImageNet1k performance. While CorrEmbed performance naturally tends to increase for each successive layer of a model, the final output layer is found to produce consistently worse embeddings than the second to last layer. We note that specific model architectures tend to surpass others with comparable performance on ImageNet. For example, Torchvision's v2 models consistently outperform their v1 counterparts. This study contributes valuable insights into the performance of various models in the context of image-embedding similarity. Though perhaps few surprising conclusions can be drawn from the results, formal evaluation compared to human annotators is an important step to ensure our RSs are as rigorous as possible.  


\section{Acknowledgments}
Funded by the Research Council of Norway through the project ``Your green, smart and endless wardrobe", project number 309977.
We thank FJONG for providing the data used as a basis for the dataset in this paper.

\bibliographystyle{ACM-Reference-Format}
\bibliography{sample-base}

\appendix

\end{document}